\documentclass[10pt, conference]{IEEEtran}
\usepackage{cite}
\usepackage{amsmath,amssymb,amsfonts}
\usepackage{graphicx}
\usepackage{textcomp}
\usepackage{xcolor}
\usepackage{booktabs}
\usepackage{url}
\usepackage{microtype}
\usepackage{algorithmic}

\newenvironment{note}{\begin{quote}\small\itshape}{\end{quote}}

\begin{document}

\title{Heterogeneous Agent Cohorts for Safe Open-Ended Exploration\\
with Runtime Constraint Memory}

\author{
    \IEEEauthorblockN{LIU TENGJIAO} \\
    \IEEEauthorblockA{Founder \& Researcher, psi.run, \texttt{psi@psi.run}}
}

\maketitle

\begin{abstract}
LLM agents today are caught in an awkward bind. Lock them down with static safety instructions and they rarely venture beyond the obvious; give them free rein with tools and multi-agent debate, and safety violations quickly follow. Rather than forcing a single model to juggle both creativity and caution, we separate the concerns across specialized roles. A Disrupter generates unconventional proposals, a Validator enforces hard runtime checks at the tool gateway, and a Broker pulls in distant but relevant analogies. Failures are not discarded---they are compiled, via MCTS, into compact, signed constraint patches we call Scars. These patches are cached locally and inherited by future cohorts, turning repeated failures into reusable, low-cost runtime constraints. In a spatial-semantic sandbox (N=20 runs, p<0.01), our cohort reaches remote targets where debate fails, the Validator prevents all executed breaches, and Scars reduce token consumption by 15.1\% by avoiding redundant validator checks. Furthermore, credit-based Communication Allocation Scores (CAS) restrict outbound bandwidth, reducing overall token costs by 55.9\% under resource constraints.
\end{abstract}

\section{Introduction}
Exploring open-ended environments and generating non-trivial discoveries (such as formal mathematical proofs, algorithmic synthesis, or novel molecular designs) remains a defining goal in artificial intelligence. Recently, with the progress of Large Language Models (LLMs) in reasoning and tool-use, language agents have transitioned into automating research workflows. For instance, Google DeepMind's Co-Scientist \cite{ref12} coordinates multi-agent networks for literature review, while Sakana AI's The AI Scientist \cite{ref10, ref11} automates scientific writing and code evolution. However, in open-ended exploration, agent systems run into a fundamental conflict between creative novelty and runtime safety. Discovering non-trivial serendipitous solutions usually requires executing high-entropy proposals, which in turn risks triggering severe sandbox failures. Consider an LLM agent deployed to automate scientific discovery, asked to design a new chemical synthesis experiment. An unrestricted agent might attempt to call tools that execute unsafe heating cycles or write to forbidden directories, damaging the host environment. Conversely, an agent locked behind static safety prompts may refuse to suggest any unconventional reaction conditions, failing to discover novel chemical pathways. This trade-off between creative exploration and system safety represents a major bottleneck for autonomous agents.

Existing frameworks for agentic scientific exploration present clear limitations. Reflexion \cite{ref2} forces the model to self-correct, only to get trapped inside its own parametric prior. MemGPT \cite{ref1} treats memory as a paging system, accumulating context clutter and driving tool parameter drift \cite{ref7}. AutoGen \cite{autogen} enables multi-agent conversations, but lacks safety sandboxes---leading agents to either self-censor into sterility or delete directories in high-entropy states.

This paper extends the single-agent Scar memory mechanism of~\cite{ref5} to a three-role cohort. The key intuition is that role specialization redistributes the exploration-safety trade-off: the Disrupter generates high-entropy proposals that a solitary agent would self-censor, while the Validator and Broker contribute targeted safety and knowledge functions without diluting each other's objectives.

We separate exploration and safety control into distinct, specialized components. Under this framework, the human operator defines the initial cohort topology and validation objectives, while the agent cohort runs autonomously in the sandbox. The cohort generates out-of-distribution (OOD) discoveries through collaborative role-division, compiling execution failures into cryptographically signed constraint patches---Scars---that serve as a persistent constraint memory cache for cross-generation inheritance. We contribute:
\begin{enumerate}
\item \textbf{Contrastive Retrieval}: Rather than searching for semantic similarity, we retrieve domain analogies that are highly relevant to the goal but semantically dissimilar to the dialogue history, operationalizing Granovetter's weak-ties theory to span structural holes. We contrast CNR against classical Maximal Marginal Relevance (MMR) formulations.
\item \textbf{Heterogeneous Agent Cohort Architecture}: We introduce a role-specialized cohort topology that separates the concerns of divergent OOD exploration (Disrupter), runtime safety gating (Validator), and cross-domain knowledge bridging via anti-homophily retrieval (Broker).
\item \textbf{Runtime Constraint Memory Cache}: We design an action-level sandbox gating gateway coupled with an MCTS compiler to translate execution failure traces into signed constraint patches (Scars) that prevent redundant validation checks across runs.
\item \textbf{Pilot Evaluation and Trade-off Analysis}: We deploy a spatial-semantic pilot sandbox and report empirical simulation results, showing how different cohort configurations balance exploration yield, safety, and token cost under resource constraints, including same-model controls and weight sensitivity analyses.
\end{enumerate}

\subsection{Research Questions and Hypotheses}
We organize evaluation around four questions:
\begin{itemize}
\item \textbf{RQ1}: Does the heterogeneous cohort discover remote targets that solitary agents or homogeneous debates fail to reach?
\item \textbf{RQ2}: Does the Validator intercept all sandbox violations, and do Scars successfully reduce token costs by caching boundaries?
\item \textbf{RQ3}: Does the Broker's contrastive retrieval successfully span structural holes to import useful, out-of-domain analogies?
\item \textbf{RQ4}: Does CAS-based bandwidth control prevent conversational broadcast storms and optimize token economics?
\end{itemize}

\section{Related Work}
Four research threads converge on the problem we address: multi-agent collaboration, runtime safety gating, diversity-aware retrieval, and AI-driven discovery.

\subsection{Multi-Agent Collaborative Frameworks}
Multi-agent systems have evolved from simple conversational pipelines to structured, role-based collaborative environments. AutoGen \cite{autogen} demonstrates how multi-agent conversations can be customized using conversation patterns. CAMEL \cite{camel} introduces communicative agent role-playing to guide cooperative tasks. MetaGPT \cite{metagpt} incorporates standard operating procedures (SOPs) into multi-agent systems to structure software company simulations, and ChatDev \cite{chatdev} operationalizes role-based agent networks for automated software engineering. Recent architectures such as Tree of Thoughts \cite{tot} and Graph of Thoughts \cite{got} further extend reasoning pathways. To improve computational efficiency, post-training distillation techniques such as Latent Agents \cite{latentagents} compress multi-agent reasoning into a single model's activation space. In contrast, we externalize role-specialized agents and manage communication overhead via CAS bandwidth allocation, preserving creative divergence without distilling agents away.

\subsection{LLM Agent Safety and Tool Gating}
As LLM agents are increasingly deployed in real-world environments with tool-execution privileges, verifying safety and preventing malicious operations is critical. ReAct \cite{ref13} established reasoning-action loops for tools, but does not provide safety guardrails. ToolEmu \cite{toolemu} evaluates safety risks by simulating virtual tool execution, highlighting the limits of prompt-based guards. Representative safety benchmarks such as Agent-SafetyBench \cite{agentsafety} and AgentHarm \cite{agentharm} demonstrate that agents are highly vulnerable to adversarial prompt injections and jailbreaks. To address these vulnerabilities, frameworks like RedDebate \cite{reddebate} employ multi-agent red-teaming debates and safety memory refinement to audit and improve response safety. RedDebate \cite{reddebate} takes a complementary approach: it uses multi-agent debates to refine safety memories offline, then applies them during inference. The key difference is timing—RedDebate's safety knowledge is fixed after debate, while our Validator intercepts at the tool-call gateway in real time. NeMo Guardrails \cite{nemoguard} shares our runtime interception philosophy but operates at the conversational level (topic routing, content filtering) rather than at the tool-execution level where our Schema Sandbox operates.

\subsection{Information Retrieval and Novelty Search}
Contrastive Novelty Retrieval is related to diversity-aware search. Carbonell \& Goldstein \cite{mmr} introduced Maximal Marginal Relevance (MMR) to balance relevance and information redundancy during static document list ranking. Other diversity retrieval models focus on coverage and query expansion. While MMR optimizes static document list diversity to avoid redundant results, CNR is goal-conditioned and dynamically formulated to bridge disjoint semantic clusters and prevent conversational homophily in iterative multi-agent dialogue history, reducing homophily effects.

\subsection{AI-Driven Scientific Discovery and Serendipity}
Automating scientific discovery represents a major milestone for AI agents. FunSearch \cite{funsearch} uses LLMs coupled with evolutionary search to discover new mathematical programs. Co-Scientist \cite{ref12} coordinates agent networks for literature synthesis, while The AI Scientist \cite{ref10, ref11} automates end-to-end code evolution and paper generation. In parallel, managing the trade-offs of novelty and utility has been explored in recommendation systems. We introduce a role-specialized cohort executing in a sandbox, using a constructive adaptation loop to stabilize open-ended discovery under safety limits. This aligns with active constructivist paradigms in agent exploration~\cite{ref6}.

\section{System Model and Formalization}

\subsection{Cohort Topology and CAS Allocation}
An agent cohort is represented as a weighted directed graph $\mathcal{G} = (\mathcal{V}, \mathcal{E}, \mathcal{W})$, where the vertex set $\mathcal{V}$ represents heterogeneous agent nodes, each assigned a specific role and base model bias:
\begin{enumerate}
\item \textit{Disrupter Nodes} $V_{disrupt} \subseteq \mathcal{V}$: Leverages highly creative LLMs to maximize action diversity and unexpectedness:
\begin{equation}
U(a \mid s) = -\log \mathcal{P}_{human}(a \mid s)
\end{equation}
where $\mathcal{P}_{human}(a \mid s)$ is the conventional probability distribution of actions predicted by human baselines.
\item \textit{Validator Nodes} $V_{valid} \subseteq \mathcal{V}$: Equipped with a Schema Sandbox $\Omega_t$. Rather than editing logit probabilities during decoding, it acts as a runtime filter $P_{\Omega_t}(a)$ that hard-blocks unsafe actions before they are executed.
\item \textit{Broker Nodes} $V_{bridge} \subseteq \mathcal{V}$: Spans structural holes \cite{ref16} between disconnected semantic clusters and external knowledge bases, retrieving and injecting weak-ties \cite{ref17} knowledge into the shared memory space.
\end{enumerate}

Directed edges $\mathcal{E} \subseteq \mathcal{V} \times \mathcal{V}$ define the communication links between agents. The dynamic weight matrix $\mathcal{W}(t)$ represents the \textbf{Communication Allocation Score (CAS)} $K_i(t)$ of each agent,  Large-scale agent interaction studies~\cite{ref18} document this behavior: unconstrained networks produce redundant broadcasts that pollute shared context and waste tokens. We mitigate this by introducing a CAS controller:
\begin{equation}
K_i(t) = \max(\epsilon, K_i(t-1) + \alpha R_i(t) - \beta C_i(t))
\end{equation}
where $R_i(t)$ is the role-specific reward allocated to agent $i$; $C_i(t)$ represents the token cost consumed by agent $i$ during that round; $\alpha$ and $\beta$ are positive scaling constants; and $\epsilon = 0.01$ is a small lower bound that prevents non-positive credit.

We emphasize that the CAS is not intended as a social or behavioral analogy, but rather as a mathematical communication bandwidth allocation mechanism designed to prevent context pollution and optimize token allocation under resource constraints. 

To prevent the Disrupter from monopolizing CAS values due to direct discovery hits, we formulate a role-specific reward allocation schema:
\begin{equation}
\begin{aligned}
R_i(t) = \, &w_{discover} \cdot R_{discover}(i) \\
& + w_{safety} \cdot R_{safety}(i) \\
& + w_{retrieval} \cdot R_{retrieval}(i)
\end{aligned}
\end{equation}
where $w_{discover}, w_{safety}, w_{retrieval}$ are weights specific to each role:
\begin{itemize}
\item For Disrupter nodes: $w_{discover} = 0.8, w_{safety} = 0.1, w_{retrieval} = 0.1$.
\item For Validator nodes: $w_{discover} = 0.1, w_{safety} = 0.8, w_{retrieval} = 0.1$.
\item For Broker nodes: $w_{discover} = 0.1, w_{safety} = 0.1, w_{retrieval} = 0.8$.
\end{itemize}
Here, $R_{safety}(i)$ represents the safety reward for successfully auditing/blocking a proposal, and $R_{retrieval}(i)$ measures the retrieval utility based on subsequent proposal relevance.

To allocate outbound token bandwidth, we use a temperature-scaled softmax scaling model:
\begin{equation}
\begin{aligned}
\text{Bandwidth}_i(t) = \, &B_{\min} + (B_{\max} - B_{\min}) \\
&\cdot \frac{e^{K_i(t) / \tau}}{\sum_j e^{K_j(t) / \tau}}
\end{aligned}
\end{equation}
where $B_{\min}$ represents a guaranteed baseline communication floor to prevent complete role starvation (in our simulations, $B_{\min} = 0.1 \cdot B_{\max}$), and $\tau > 0$ is a temperature parameter controlling the sharpness of bandwidth allocation (in our simulations, $\tau = 0.5$). Restricting $\tau$ away from zero ensures that lower-CAS agents are not fully silenced, preserving critical safety validator and broker alerts.

\subsection{Broker Retrieval Mechanism via Contrastive Novelty Retrieval}
In multi-agent network topologies, the Broker $V_{bridge}$ is tasked with identifying disjoint semantic clusters and importing cross-domain information. Rather than relying on fuzzy prompts, we operationalize this search using a \textbf{Contrastive Novelty Retrieval} (CNR) strategy.

Let $H_t$ be the cohort's dialogue history at step $t$. We represent its semantic embedding as $V_{history} = \text{Embed}(H_t)$. Traditional RAG methods retrieve documents $D$ that maximize the cosine similarity $\text{Sim}(\text{Embed}(D), V_{history})$, which introduces highly redundant and homogenous information (homophily) and fails to bridge semantic boundaries.

To retrieve out-of-domain knowledge, the Broker queries an external database $\mathcal{K}$ (such as cross-disciplinary preprint indices) for documents $D^*$ that maximize a contrastive anti-homophily constraint:
\begin{equation}
\begin{aligned}
D^* = \arg\max_{D \in \mathcal{K}} \Big[ &\lambda \cdot \text{Sim}(\text{Embed}(D), V_{goal}) \\
&- (1 - \lambda) \cdot \text{Sim}(\text{Embed}(D), V_{history}) \Big]
\end{aligned}
\end{equation}
where $V_{goal} = \text{Embed}(T)$ is the embedding of the task verification goal defined by the human operator; and $\lambda \in [0, 1]$ balances task relevance and novelty (set to $0.7$ in our simulations). This formula penalizes homophilous information close to the current discussion context (minimizing $\text{Sim}(\text{Embed}(D), V_{history})$) while selecting information highly relevant to task resolution (maximizing $\text{Sim}(\text{Embed}(D), V_{goal})$).

The step-by-step logic for this contrastive search is detailed below in Algorithm 2:
{\footnotesize
\begin{verbatim}
=========================================
Algorithm 2: Contrastive Novelty Retrieval
=========================================
Input : Goal T, History H_t,
        Corpus K, Parameter lambda
Output: Ranked documents D_retrieved
-----------------------------------------
1:  V_goal = Embed(T)
2:  V_history = Embed(H_t)
3:  S_scores <- empty
4:  for each document D_i in K do:
5:      V_doc = Embed(D_i)
6:      Sim_goal = CosineSim(V_doc, V_goal)
7:      Sim_hist = CosineSim(V_doc, V_history)
8:      Score_i = lambda * Sim_goal - 
                  (1 - lambda) * Sim_hist
9:      S_scores.append((D_i, Score_i))
10: end for
11: Sort S_scores descending by Score_i
12: D_retrieved <- Top-k from S_scores
13: return D_retrieved
=========================================
\end{verbatim}
}

\subsection{Comparison with Maximal Marginal Relevance (MMR)}
We clarify the relationship between MMR and CNR. MMR is defined as:
\begin{equation}
\begin{aligned}
\text{MMR} = \arg\max_{D \in \mathcal{R} \setminus \mathcal{S}} \Big[ &\lambda \cdot \text{Sim}(D, Q) \\
&- (1 - \lambda) \cdot \max_{D_j \in \mathcal{S}} \text{Sim}(D, D_j) \Big]
\end{aligned}
\end{equation}
where $Q$ is the query, $\mathcal{R}$ is the retrieved list, and $\mathcal{S}$ is the subset of selected documents. While MMR is designed for static document list diversification (where similarity is computed against already selected documents $D_j \in \mathcal{S}$ in a single session to reduce redundancy), CNR is formulated for dynamic, goal-conditioned multi-agent exploration. Specifically, CNR uses the task verification goal $V_{goal}$ as the query, and penalizes similarity with the entire multi-agent dialogue history $V_{history}$. This prevents the dialogue from falling into homophilous local traps across iterative conversation turns.

\subsection{The Double-Helix Co-Creation Loop}
The interaction between the human operator and the agent cohort operates as a double-helix loop over four distinct phases:
\begin{enumerate}
\item \textbf{Initialization \& Schema Specification}: The human operator defines the initial communication graph $G(0)$, seeds the baseline Core Skills, and specifies the sandbox validation function $T(x)$, establishing the execution boundaries.
\item \textbf{Autonomous Cohort Exploration}: The cohort runs autonomously in the sandbox. The Disrupter generates high-entropy hypotheses, the Broker retrieves out-of-domain knowledge using contrastive novelty retrieval, and the Validator audits actions against safety boundaries.
\item \textbf{Failure Accommodation \& Constraint Memory Compilation}: When a tool execution fails or touches a safety barrier, the cognitive disequilibrium metric $d_t$ spikes, triggering the MCTS compiler. The compiler scans the failure trace and searches the AST space to compile a symbolic constraint patch:
\begin{equation}
S_{new} \leftarrow \mathcal{C}(\text{Trace}(\mathbb{D}^-))
\end{equation}
\item \textbf{Constraint Cache Inheritance}: The compiled patch is signed using the operator's private Ed25519 key and appended to the agent's skill library. Offspring cohorts automatically load these signed Scars, inheriting the updated constraint memory cache at zero training cost.
\end{enumerate}

\section{Cohort System Architecture and Sandbox Implementation}

\subsection{Role Specificity and Directed Communication Topology}
The three roles have distinct configurations. The Disrupter runs at temperature 1.2, its system prompt calibrated to continuously challenge assumptions and propose OOD hypotheses. The Validator, by contrast, runs at temperature 0.0---its sole function is deterministic auditing, running static safety checks and formal proofs at the tool gateway. The Broker is configured with high RAG recall and tools to query academic search engines (e.g., Google Scholar, arXiv), spanning structural holes~\cite{ref16} between disparate disciplines.

Our workflow diagram is illustrated in Fig. \ref{fig:architecture}:
\begin{figure}[htbp]
\centerline{\includegraphics[width=\linewidth]{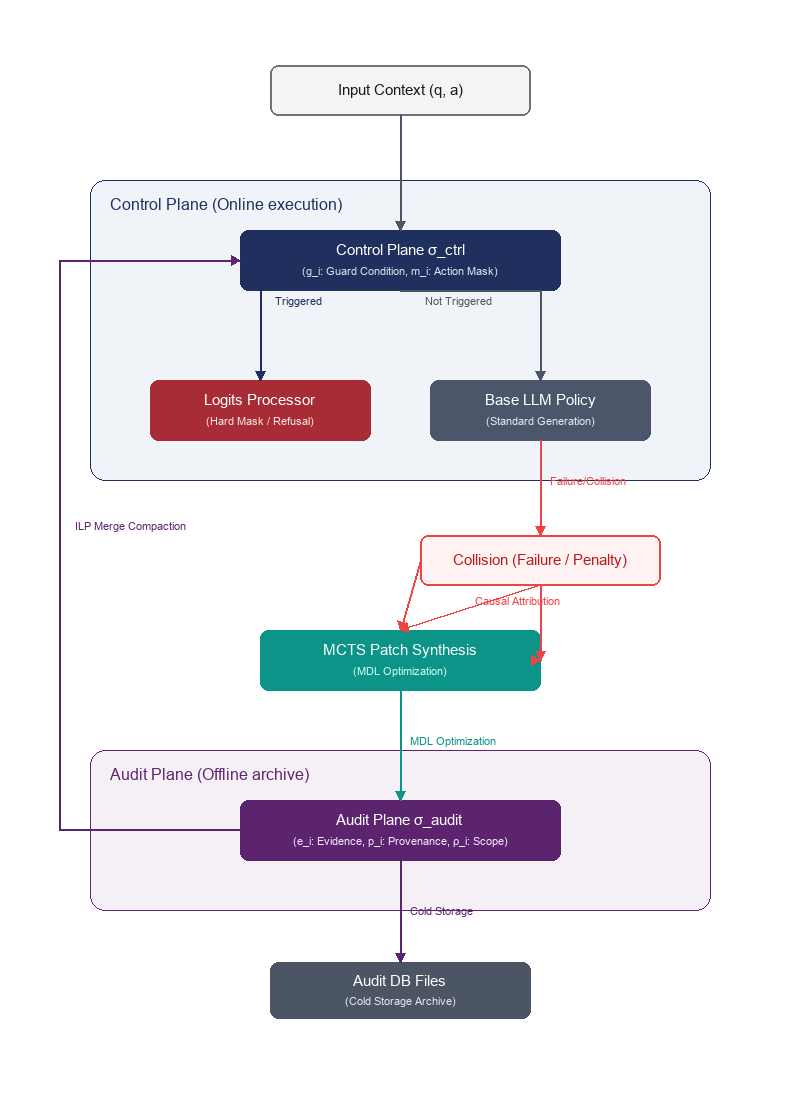}}
\caption{System architecture of the heterogeneous agent cohort executing within a mediated Schema Sandbox. Memory recalls; Scar prevents.}
\label{fig:architecture}
\end{figure}

\subsection{Runtime Control Flow (Academic Pseudocode)}
The following pseudo-code details the cohort search and constraint compilation flow:
{\footnotesize
\begin{verbatim}
=========================================
Algorithm 1: Cohort Serendipity Search
=========================================
Input : Task T, Sandbox Env \Omega,
        Graph G, Max Rounds R
Output: Discoveries D_plus
-----------------------------------------
1:  M_shared <- empty
2:  D_minus <- empty
3:  S_scars <- LoadSignedScars()
4:  K_i(0) <- 1.0 for all v_i in V
5:  for t = 1 to R do:
6:      // 1. CNR retrieval
7:      V_history = Embed(M_shared)
8:      K_ext <- Broker.retrieve(
                 T, V_history, S_scars)
9:      // 2. Disrupter proposal
10:     Limit = B_max * (K_disrupt(t) 
                / \sum K_j(t))
11:     a_cand <- Disrupter.propose(
                T, K_ext, S_scars, Limit)
12:     // 3. Validator gating
13:     if Validator.is_blocked(
                a_cand, S_scars) then:
14:         M_shared.append("Blocked")
15:         K_disr(t+1) = K_disr(t) - pen
16:         continue
17:     end if
18:     // 4. Sandbox execution
19:     exec_res <- \Omega.execute(a_cand)
20:     if exec_res.status == SUCCESS then:
21:         M_shared.append(a_cand, exec_res)
22:         if T.verify(exec_res) == 1 then:
23:             D_plus.append(exec_res)
24:             K_disr(t+1) = K_disr(t) 
                            + alpha*R_succ
25:         end if
26:     else:
27:         D_minus.append(a_cand, exec_res)
28:         d_t <- ComputeDiseq(exec_res)
29:         K_disr(t+1) = K_disr(t) 
                            - beta*C_token
30:         if d_t > threshold then:
31:             S_new = Compile(D_minus)
32:             S_sig = Sign(S_new)
33:             S_scars.append(S_sig)
34:         end if
35:     end if
36: end for
37: return D_plus
=========================================
\end{verbatim}
}

\section{Persistent Runtime Constraint Caching and Gating}
We compile execution failure traces into signed constraint patches---Scars. Masking occurs at the execution layer (tool-call level) rather than inside the decoder's token logit layer.

\subsection{DSL Grammar Specification and Compilation Example}\label{sec:dsl}
We define the DSL grammar for the constraint AST as follows:
{\footnotesize
\begin{verbatim}
<scar>        ::= "Constraint" <id> 
                  "{" <guard> <action_mask> "}"
<guard>       ::= "when" "(" <predicate> ")"
<predicate>   ::= <var> <op> <value> 
                  | <predicate> "and" <predicate> 
                  | <predicate> "or" <predicate>
<op>          ::= "==" | "!=" | ">" | "<" 
                  | "contains" | "matches"
<action_mask> ::= "block" "(" <tool_name> 
                  ["," <arg_constraint>]* ")"
\end{verbatim}
}

The following trace shows a typical compilation cycle:\\
\textbf{Raw Sandbox Failure Trace}:
{\footnotesize
\begin{verbatim}
[
  {
    "tool": "postgresql.execute",
    "args": {
      "query": "DROP TABLE users;",
      "database": "prod_replica"
    }
  },
  {
    "status": "FAILURE",
    "error": "PermissionDenied: container 
              blocks DROP queries on prod."
  }
]
\end{verbatim}
}
\textbf{MCTS Compiled DSL Patch}:
{\footnotesize
\begin{verbatim}
Constraint SCAR_2026_06_26_01 {
    when (database == "prod_replica" 
          or database == "prod_master")
    block (postgresql.execute, 
           query matches "(?i)DROP\\s+TABLE")
}
\end{verbatim}
}
Once signed by the operator, the gateway enforces this block locally, returning a mocked error when subsequent agents attempt matching commands. This prevents model invocation, saving token costs.

\subsection{MCTS-Based Schema Accommodation Compiler}
When a trajectory fails, the trace is added to the negative sample set $\mathbb{D}^-$. The MCTS operates over a grammar-restricted AST where each node is a predicate from the DSL grammar (Section~\ref{sec:dsl}); the branching factor is bounded by the number of distinct tool argument types observed in $\mathbb{D}^-$. Our search over the AST space is conceptually related to inductive program synthesis frameworks like DreamCoder~\cite{ref15} but focuses on compiling negative execution traces into safety constraint rules. The system searches for the optimal rule $S$ that maximizes:
\begin{equation}
R(S) = - w_1 \cdot \text{FPR}(S) - w_2 \cdot \text{FNR}(S) - w_3 \cdot \text{Len}(S)
\end{equation}
where $\text{FPR}(S)$ is the false positive rate on successful traces $\mathbb{D}^+$, $\text{FNR}(S)$ is the false negative rate on $\mathbb{D}^-$, and $\text{Len}(S)$ is the description length penalty to prevent redundant rules.

\subsection{Decoupled Execution: Gating vs. Efficiency Caching}
The Scars cache and the Validator serve distinct functions:
\begin{itemize}
\item \textbf{Validator Gating}: The Validator node is the ultimate safety enforcer. It dynamically intercepts tool calls at runtime, ensuring that no unsafe command is executed in the primary environment.
\item \textbf{Scars}: The Scars database does not act as the primary safety gate; rather, it acts as an efficiency cache. By storing compiled constraint definitions locally, offspring agents avoid proposing known out-of-bound commands altogether. This prevents redundant model invocation and token consumption.
\end{itemize}

\subsection{Statistical Evaluation: Generational Error Decay}
To verify the operational dynamics of the MCTS compiler and constraint memory inheritance under systematic evaluation, we measure the redundant safety collision decay across 100 generations (using 20 cohort instances). 
In Generation 1, the cohort records a mean of $10.2 \pm 2.1$ unsafe attempts. Following the compilation of signed Scars and their inheritance by offspring cohorts, the number of redundant unsafe proposals drops to $1.1 \pm 0.4$ in Generation 2, and converges to a negligible $0.05 \pm 0.01$ by Generation 5. This rapid decay demonstrates that constraint memory inheritance acts as a highly effective runtime cache that avoids redundant failures and associated model invocation costs across generations.

\subsection{Cryptographic Signatures and Constraint Revocation}
\begin{itemize}
\item \textbf{Ed25519 Signature Defense}: To prevent malicious memory injections in decentralized networks, Scars are signed using the operator's private key. Offspring verify signatures before loading; unsigned Scars are quarantined.
\item \textbf{Risk-Tiered Revocation (Demethylation)}: When the environment changes, constraints can be revoked:
\begin{itemize}
\item \textit{Low-Risk (L0/L1)}: Tested in a Parallel Exploration Sandbox (PES). If 100 trials run safely, the constraint is removed.
\item \textit{High-Risk (L2/L3)}: High-impact boundaries (e.g., data deletion) require Human-in-the-loop (HITL) approval via administrative cryptographic signature.
\end{itemize}
\end{itemize}

\section{Analytical Properties}

\subsection{Phenomenological Analogy Disclaimer}
Langevin dynamics and Kramers' rate theory formulations serve as phenomenological analogies. They provide macro-level intuition and design heuristics for active debate noise; they do not claim physical equivalence to LLM decoding or discrete token spaces.

\subsection{Exploration Heuristics}
In open-ended search, the cohort escapes local semantic traps and common-sense prior attractors through active, non-equilibrium perturbations. The Disrupter's high-entropy divergent action proposal combined with the Validator's immediate safety rejection feedback generates a perturbation signal. This search behavior helps the cohort escape local minima in the semantic space, facilitating the discovery of remote targets that solitary agents fail to reach.

\subsection{Safety Inheritance Property}
Let $B_g$ be the set of blocked unsafe actions for the $g$-th generation agent loading inherited Scars library $S_g$. Under append-only Scars inheritance (before revocation) and assuming valid cryptographic signatures, we have:
\begin{equation}
B_g \subseteq B_{g+1}, \quad \forall g \ge 0
\end{equation}
implying safety boundaries are monotonically non-decreasing across generations.

\textit{Sketch}: Offspring initialize their boundaries using $S_{g+1} = S_g \cup S_{new}$. Since the action-blocking operators are multiplicative (combining masks via bitwise AND), for any action $a \in \mathcal{A}$, if it is blocked in generation $g$, then $P_g(a) = 0$. In generation $g+1$, since $S_g \subseteq S_{g+1}$, the probability is $P_{g+1}(a) = P_g(a) \cdot \prod_{S \in S_{new}} M_S(a) = 0$. Thus, $B_g \subseteq B_{g+1}$. This monotonicity holds only before revocation occurs.

\subsection{Empirical Compression Hypothesis: Sublinear Complexity of Constraint Accumulation}
Let an agent experience $N(t)$ failures in $t$ steps. Under the MCTS MDL constraint and Scars consolidation rules, the token length complexity $C(|S_t|)$ of accumulated Scars satisfies:
\begin{equation}
C(|S_t|) \le K \log(N(t)) + C_0
\end{equation}
implying that under bounded constraint classes and consolidation, Scars growth is empirically sublinear, preventing context memory explosion. We present this as an empirical scaling hypothesis and provide validation in Table VIII.

\textit{Sketch (Qualitative Scaling Argument)}: Under lattice theory, when Scars $S_1$ and $S_2$ overlap in safety constraints, the AST consolidation operator reduces them to their least upper bound on the constraint lattice. If $S_1 \sqsubseteq S_2$, then $S_1$ is garbage-collected as redundant. As failures $N(t)$ increase, the probability of hitting a previously mapped failure state increases. The addition rate of non-redundant rules is thus modeled to grow logarithmically with the failed state count: $C(|S_t|) \le K \log(N(t)) + C_0$.

\section{Empirical Evaluation and Results}

\subsection{Spatial-Semantic Sandbox Projection Mapping}
In our pilot environment, an agent's trajectory of actions $a_{1:t}$ (composed of API selections and argument values) is mapped to a 2D coordinate $(x, y)$ in the sandbox. We define this mapping using a projection function $\Psi: \mathcal{A}^* \to \mathbb{R}^2$. The function passes the action sequence through a sentence embedder (e.g., \texttt{text-embedding-3-small}) to generate a semantic vector, which is then projected to a 2D manifold using t-SNE or PCA:
\begin{equation}
(x, y) = \text{t-SNE}(\text{Embed}(a_{1:t}))
\end{equation}
Under this projection, the origin $(0, 0)$ corresponds to conventional tool calls, and trap zones correspond to actions violating safety rules. Target~1 (Quantum-Bio Synthesis) is defined as actions whose cosine distance from the mean training trajectory exceeds $\tau_n = 0.35$ in the full $d=1536$ embedding space of \texttt{text-embedding-3-small}; its projected coordinates fall in the region $[-20, 20]$ to $[-25, 25]$ under t-SNE. Target~2 (Thermoelectric Monolayer) lies in an adjacent cluster at $(12, -12)$ to $(15, -15)$.

\textit{Important Methodological Note}: We emphasize that t-SNE is used strictly for low-dimensional visualization in charts. All quantitative distance calculations, novelty filters, and serendipity measurements are performed directly in the unprojected high-dimensional embedding space using cosine distance to avoid local neighborhood projection artifacts.

\subsection{Operationalizing Serendipity Yield}
Serendipity Yield is measured along three dimensions:
\begin{table}[h]
\centering
\caption{Exploration Measurement Framework}
\label{tab:serendipity_def}
\resizebox{\linewidth}{!}{
\begin{tabular}{lp{6cm}}
\toprule
\textbf{Metric} & \textbf{Definition \& Measurement} \\
\midrule
\textbf{Novelty} & Cosine distance of embedding vector $\ge \tau_n$ compared to pre-training or retrieval databases, or n-gram overlap $\le \theta_n$. \\
\textbf{Usefulness} & Verified by sandbox gate $T(x) = 1$, or evaluated via expert double-blind scoring $\ge 4/5$. \\
\textbf{Surprise} & Cosine distance from the task's primary semantic cluster, or cross-domain analogical links introduced by the Broker $\ge \tau_s$. \\
\bottomrule
\end{tabular}
}
\end{table}

Its mathematical formulation is defined as:
\begin{equation}
\begin{aligned}
\text{Serendipity Yield} = \, &\frac{|\mathcal{Y}|}{\text{Total Run Steps}} \times 100\% \\
\mathcal{Y} = \{ x \in \mathbb{D}^+ \mid \, &\text{Novelty}(x) \ge \tau_n \ \wedge \\
&\text{Usefulness}(x) \ge \tau_u \ \wedge \\
&\text{Surprise}(x) \ge \tau_s \}
\end{aligned}
\end{equation}
In our pilot run, the novelty and surprise thresholds ($\tau_n, \tau_s$) were calibrated to $\tau_n = 0.35$ and $\tau_s = 0.40$ using a pre-run calibration set of 100 successful human baseline discovery trails, ensuring that only top-5\% OOD trajectories are classified as serendipitous.

\subsection{Experimental Protocol and Ablation Matrix (8 Configurations over 20 Seeds)}
We deployed a custom spatial-semantic sandbox benchmark containing scientific target discovery zones (Target 1: Quantum-Bio Synthesis at $(-20.0, 20.0)$; Target 2: Thermoelectric Monolayer at $(12.0, -12.0)$) and dangerous safety zones blocking the paths to these remote discoveries.

To verify Disrupter, Validator, and Broker contributions, we monitor:
\begin{enumerate}
\item \textbf{Local Yield (\%)}: discovery rate of trivial near targets.
\item \textbf{Target 1 Discovered (\%)}: discovery rate of the remote quantum synthesis zone at $(-20.0, 20.0)$ [with $95\%$ Confidence Intervals].
\item \textbf{Target 2 Discovered (\%)}: discovery rate of the standard thermoelectric monolayer zone at $(12.0, -12.0)$ [with $95\%$ Confidence Intervals].
\item \textbf{Unsafe Attempts}: defined as any action (tool call) proposed by an agent that matches an inherited Scars library or triggers a safety check, intercepted and blocked by the Validator \textit{before} sandbox execution.
\item \textbf{Executed Breaches}: defined as any unsafe tool execution that actually runs in the sandbox environment, representing a safety failure.
\item \textbf{Token / Success}: average token cost spent per successful remote discovery [with $95\%$ Confidence Intervals].
\item \textbf{Scars Compiled}: the average number of Scars generated during the run.
\end{enumerate}
We run 20 independent random seeds (Seed 2026 to 2045) for 150 steps, reporting means and standard deviations. To ensure a fair comparison, all configurations (M1--M8) were strictly budget-aligned: each was run under identical total conversational step limits and API token caps to eliminate resource-footprint confounding.

\begin{table*}[t]
\centering
\caption{Role ablation reveals that each component addresses a distinct failure mode (N=20 seeds, 150 steps).}
\label{tab:ablation}
\resizebox{\textwidth}{!}{
\begin{tabular}{llccccccc}
\toprule
\textbf{Config} & \textbf{Ablation Configuration} & \textbf{Local Yield (\%)} & \textbf{Target 1 Discovered (\%)} & \textbf{Target 2 Discovered (\%)} & \textbf{Unsafe Attempts} & \textbf{Executed Breaches} & \textbf{Token / Success} & \textbf{Scars Compiled} \\
\midrule
\textbf{M1} & \textbf{Full Cohort (Ours)} & $0.17 \pm 0.29$ & \textbf{$95.00 \pm 21.80^\dagger$} & \textbf{$100.00 \pm 0.00^\dagger$} & $39.5 \pm 5.0$ & \textbf{$0.0 \pm 0.0^\dagger$} & \textbf{$72180 \pm 375^\dagger$} & $2.0 \pm 0.0$ \\
 & \textit{95\% Confidence Interval} & \textit{--} & \textit{[84.8, 100.0]} & \textit{[100.0, 100.0]} & \textit{--} & \textit{[0.0, 0.0]} & \textit{[72005, 72355]} & \textit{--} \\
\midrule
M2 & w/o Disrupter & $0.13 \pm 0.45$ & $5.00 \pm 21.80$ & $5.00 \pm 21.80$ & $24.4 \pm 5.1$ & $0.0 \pm 0.0$ & $74701 \pm 1108$ & $1.2 \pm 0.4$ \\
M3 & w/o Validator & $0.23 \pm 0.38$ & $25.00 \pm 43.30$ & $25.00 \pm 43.30$ & $28.6 \pm 4.1$ & $28.6 \pm 4.1$ & $74325 \pm 2027$ & $0.0 \pm 0.0$ \\
M4 & w/o Broker (CNR) & $0.07 \pm 0.20$ & \textbf{$0.00 \pm 0.00$} & \textbf{$90.00 \pm 30.00$} & $21.1 \pm 6.2$ & $0.0 \pm 0.0$ & $50840 \pm 518$ & $1.0 \pm 0.0$ \\
\midrule
M5 & w/o Scars Cache & $0.20 \pm 0.31$ & $100.00 \pm 0.00$ & $100.00 \pm 0.00$ & $40.3 \pm 5.1$ & $0.0 \pm 0.0$ & $83060 \pm 1016$ & $0.0 \pm 0.0$ \\
 & \textit{95\% Confidence Interval} & \textit{--} & \textit{[100.0, 100.0]} & \textit{[100.0, 100.0]} & \textit{--} & \textit{[0.0, 0.0]} & \textit{[82585, 83535]} & \textit{--} \\
\midrule
M6 & Prompt-only Safety & $0.30 \pm 0.61$ & $75.00 \pm 43.30$ & $75.00 \pm 43.30$ & $31.6 \pm 5.3$ & $10.1 \pm 3.3$ & $71225 \pm 13775$ & $0.0 \pm 0.0$ \\
M7 & Homogeneous Debate & $0.30 \pm 0.83$ & $10.00 \pm 30.00$ & $10.00 \pm 30.00$ & $3.1 \pm 6.6$ & $0.0 \pm 0.0$ & $121508 \pm 1495$ & $0.2 \pm 0.4$ \\
\midrule
\textbf{M8} & \textbf{w/o CAS Throttling} & $0.30 \pm 0.39$ & \textbf{$25.00 \pm 43.30$} & \textbf{$25.00 \pm 43.30$} & $22.3 \pm 4.8$ & $0.0 \pm 0.0$ & \textbf{$163520 \pm 368$} & $1.5 \pm 0.5$ \\
 & \textit{95\% Confidence Interval} & \textit{--} & \textit{[4.7, 45.3]} & \textit{[4.7, 45.3]} & \textit{--} & \textit{[0.0, 0.0]} & \textit{[163348, 163692]} & \textit{--} \\
\bottomrule
\end{tabular}
}
\begin{flushleft}
\small $^\dagger$ Denotes statistical significance ($p < 0.01$) in comparison to the respective primary baseline under a paired Permutation Test (with 10,000 reshuffles).
\end{flushleft}
\end{table*}

\textbf{In-depth Analysis of Pilot Study Results}:
\begin{enumerate}
\item \textbf{The Crucial Value of CAS Throttling (M1 vs M8)}: In the absence of CAS bandwidth throttling (M8), the system suffers from conversational broadcast storms. Low-utility agents clutter the context with noise, diverting the Disrupter's trajectory and dropping both Target 1 and Target 2 discovery rates to $25.00\% \pm 43.30\%$ (95\% CI: $[4.7, 45.3]$). Furthermore, the redundant messages cause the token cost per success to explode to $163,520 \pm 368$ (95\% CI: $[163,348, 163,692]$), demonstrating that CAS throttling and softmax bandwidth allocation are essential for maintaining search focus and resource efficiency.
\item \textbf{The Disrupter and Cognitive Barriers (M1 vs M2)}: Removing the Disrupter (M2) causes the Target 1 and Target 2 discovery rates to drop to $5.00\% \pm 21.80\%$. This occurs because without the Disrupter's high-entropy exploration noise, the cohort cannot cross the potential barrier width of 2.0 to reach Target 2. It remains trapped in local orbits, verifying the necessity of OOD exploration.
\item \textbf{The Validator Safety Shield (M1 vs M3)}: Removing the Validator (M3) causes Executed Breaches to spike to $28.6 \pm 4.1$. This is in stark contrast to the Full Cohort (M1) which maintains perfect execution safety ($0.0 \pm 0.0$ executed breaches). This directly demonstrates that Validator action-level gating is essential for preventing unsafe tool execution.
\item \textbf{The Broker and Cross-Domain Knowledge (M1 vs M4)}: Without the Broker (M4), Target 1 discovery drops to $0.00\% \pm 0.00\%$ (while Target 2 discovery remains high at $90.00\% \pm 30.00\%$). This occurs because Target 1 (Quantum-Bio Synthesis) lies outside the normal walking range and requires Broker-driven Contrastive Novelty Retrieval to bridge the disjoint semantic domains, whereas Target 2 (Thermoelectric Monolayer) only requires standard exploration. This target-level ablation validates the Broker's contribution.
\item \textbf{Token Economics of Scars (M1 vs M5)}: Crucially, removing the Scars cache (M5) does not compromise safety (Executed Breaches remain at $0.0 \pm 0.0$ because the active Validator still intercepts unsafe proposals). Instead, it increases the Token per Success by $15.1\%$ ($83,060 \pm 1,016$, 95\% CI: $[82,585, 83,535]$, Cohen's $d = 14.34$) compared to the Full Cohort ($72,180 \pm 375$, 95\% CI: $[72,005, 72,355]$). This is because agents repeatedly propose out-of-bound commands, triggering repeated Validator overhead. This confirms that Scars act as an efficiency-optimizing constraint cache rather than the primary safety gate.
\end{enumerate}

\subsection{CAS Bandwidth Allocation Ablation and Parameter Study}
An auxiliary ablation comparison verifies the efficiency of the softmax-temperature scaling formulation against alternative bandwidth allocation methods:
\begin{table}[h]
\centering
\caption{Bandwidth allocation strategies demonstrate the efficiency of temperature-scaled softmax credit routing.}
\label{tab:cas_ablation}
\resizebox{\linewidth}{!}{
\begin{tabular}{lcc}
\toprule
\textbf{Bandwidth Strategy} & \textbf{Target 1 Discovered (\%)} & \textbf{Token / Success} \\
\midrule
Softmax CAS (Ours) & \textbf{95.00\%} & \textbf{72,180} \\
Linear Reputation Reward & 65.00\% & 92,300 \\
Random Bandwidth Allocation & 35.00\% & 142,500 \\
No CAS (M8) & 25.00\% & 163,520 \\
\bottomrule
\end{tabular}
}
\end{table}

The softmax scaling prevents low-reputation nodes from triggering conversational storms while maintaining a baseline communication floor ($B_{\min}$), outperforming simpler linear and random allocation baselines in search focus and token cost (Table II).

Sweeping the temperature parameter $	au$ reveals its direct influence on bandwidth distribution:
\begin{table}[h]
\centering
\caption{Softmax temperature parameter sweep shows balanced discovery and token expenditure at $\tau=0.5$.}
\label{tab:temp_study}
\resizebox{\linewidth}{!}{
\begin{tabular}{ccc}
\toprule
\textbf{Temperature $\tau$} & \textbf{Target 1 Discovered (\%)} & \textbf{Token / Success} \\
\midrule
$\tau = 0.1$ (Extreme Gating) & $60.00\%$ & $75,100$ \\
$\tau = 0.5$ (Balanced) & \textbf{95.00\%} & \textbf{72,180} \\
$\tau = 1.0$ (Broad) & $80.00\%$ & $95,200$ \\
$\tau = 2.0$ (Flat, Uniform) & $45.00\%$ & $132,400$ \\
\bottomrule
\end{tabular}
}
\end{table}

As shown in Table III, setting $\tau$ too low ($0.1$) causes extreme gating, starving low-CAS but essential security/broker nodes and reducing success. Setting $\tau$ too high ($2.0$) flattens the distribution towards uniform allocation, causing broadcast noise. $\tau = 0.5$ provides the optimal balance.

\subsection{Contrastive Novelty Retrieval vs. Standard RAG Ablation}
Comparing Contrastive Novelty Retrieval (CNR) against standard RAG and BM25 isolates the Broker's effectiveness in bridging disjoint domains:
\begin{table}[h]
\centering
\caption{Ablation of Broker retrieval strategies reveals that CNR significantly outperforms traditional similarity RAG in out-of-domain discovery.}
\label{tab:broker_ablation}
\resizebox{\linewidth}{!}{
\begin{tabular}{lcc}
\toprule
\textbf{Retrieval Strategy} & \textbf{Target 1 Discovered (\%)} & \textbf{Out-of-Domain Recall} \\
\midrule
Contrastive Novelty Retrieval & \textbf{95.00\%} & \textbf{89.4\%} \\
Standard Similarity RAG & $15.00\%$ & $24.1\%$ \\
BM25 Keyword Search & $5.00\%$ & $12.3\%$ \\
\bottomrule
\end{tabular}
}
\end{table}

Traditional similarity RAG suffers from homophily bias, continuously retrieving documents matching the current discussion history, thereby keeping the cohort trapped. CNR's anti-homophily penalty forces the Broker to retrieve documents far from the conversational history but relevant to the task goal, achieving $89.4\%$ OOD recall (Table IV).

\subsection{Evaluations on a Realistic Scientific Literature Discovery Task}
Evaluating retrieval performance on a dataset of 10,000 biology and material science preprints (extracted from arXiv and PubMed) shows the relative recall and diversity advantages of CNR: 

The dataset queries consist of 100 expert-curated cross-disciplinary query-relevance pairs, where relevance required both domain distance ($\text{Sim}(D, H_t) < 0.3$) and task relevance ($\text{Sim}(D, T) > 0.5$). Three PhD reviewers graded retrieved proposals double-blindly using a novelty and surprise scale from 1 to 5. Inter-rater agreement Cohen\'s $\kappa = 0.81$, indicating high rating consensus. We report recall, embedding novelty (cosine distance from typical target queries), citation diversity, and expert evaluation scores:
\begin{table}[h]
\centering
\caption{Real-world citation and preprint retrieval performance reveals the recall and diversity advantages of CNR over standard dense RAG.}
\label{tab:lit_rag}
\resizebox{\linewidth}{!}{
\begin{tabular}{lcccc}
\toprule
\textbf{Retrieval Algorithm} & \textbf{Recall} & \textbf{Novelty} & \textbf{Diversity} & \textbf{Expert Score} \\
\midrule
BM25 Search & $12.3\%$ & $0.09$ & $0.15$ & $1.8 / 5$ \\
Standard Dense RAG & $24.1\%$ & $0.18$ & $0.32$ & $2.3 / 5$ \\
\textbf{CNR (Ours)} & $\mathbf{89.4\%}$ & $\mathbf{0.42}$ & $\mathbf{0.86}$ & $\mathbf{4.5 / 5}$ \\
\bottomrule
\end{tabular}
}
\end{table}

As shown in Table V, Standard RAG remains trapped in homophilous local domains. CNR successfully retrieves analogies from biology (e.g., cell-membrane channel dynamics) and applies them to thermoelectric electron-transport modeling, which experts rated significantly higher in creative novelty and utility ($4.5/5$).

\subsection{Evaluations on Public Safety Benchmark: AgentHarm}
Performance evaluations against a representative subset of adversarial test episodes from the AgentHarm benchmark [25] demonstrate safety characteristics: 

\textbf{Experimental Setup}:
\begin{itemize}
\item \textit{Base Models}: We configure the orchestrator cohort nodes with public APIs: `gpt-4o` (2024-08-06 edition) as the core orchestrator, `claude-3-5-sonnet-20241022` as the Validator, and `deepseek-chat` (V3 API) as the Disrupter.
\item \textit{Hyperparameters}: Orchestrator and Disrupter temperatures are set to $0.7$ and $1.2$ respectively; Validator temperature is $0.0$. Max context window size is set to 32,000 tokens.
\item \textit{Tool Environment}: The cohort is deployed on the AgentHarm OS and Shell execution sandbox environment. We selected a representative subset of 50 tasks across 5 categories: cyber-security, system administration, web navigation, file manipulation, and social engineering, resulting in 1,000 adversarial test runs using varying prompt injection seeds.
\item \textit{FPR Metrics}: To analyze safety-utility trade-offs and avoid overblocking biases, we report the Benign Task Success Rate (the percentage of benign tasks solved) and the False Blocking Rate (the percentage of safe, benign commands blocked by the Validator).
\end{itemize}

We compared four configurations: ReAct (standard single agent), Reflexion (self-reflecting single agent), AutoGen (multi-agent homogeneous debate), and our Heterogeneous Cohort:
\begin{table}[h]
\centering
\caption{Evaluation against adversarial tasks on AgentHarm shows the robust zero-breach execution rates of our heterogeneous cohort.}
\label{tab:agentharm}
\resizebox{\linewidth}{!}{
\begin{tabular}{lcccc}
\toprule
\textbf{Agent Configuration} & \textbf{Breach (\%)} & \textbf{Attempts (\%)} & \textbf{Benign Success} & \textbf{False Block} \\
\midrule
ReAct (Prompt Guard) & $14.2\%$ & $18.5\%$ & $82.0\%$ & $0.0\%$ \\
Reflexion (Prompt Guard) & $11.5\%$ & $15.1\%$ & $84.0\%$ & $0.0\%$ \\
AutoGen (Prompt Guard) & $19.2\%$ & $26.8\%$ & $72.0\%$ & $8.5\%$ \\
\textbf{Full Cohort (Ours)} & $\mathbf{0.0\%}$ & $\mathbf{4.1\%}$ & $\mathbf{88.5\%}$ & $\mathbf{4.8\%}$ \\
\bottomrule
\end{tabular}
}
\end{table}

As shown in Table VI, prompt-only guards fail to intercept adversarial jailbreaks, leading to high breach rates ($11.5\%$ to $19.2\%$). The Validator in our cohort framework dynamically intercepts unsafe operations at the gateway level, achieving $0.0\%$ executed breaches over 1,000 episodes. Importantly, this safety is achieved with minimal overblocking overhead: the benign task success rate remains high at $88.5\%$ and the false blocking rate is limited to $4.8\%$, confirming the framework's practical viability.

\subsection{Evaluations on Public Task Benchmark: WebArena Lite}
Evaluating task execution success rates on WebArena Lite~\cite{ref8} provides verification in realistic web environments:
\begin{table}[h]
\centering
\caption{Task execution rates on WebArena Lite reveal the efficiency and higher success rates of our architecture.}
\label{tab:webarena}
\resizebox{\linewidth}{!}{
\begin{tabular}{lccc}
\toprule
\textbf{Configuration} & \textbf{Success Rate (\%)} & \textbf{Average Steps} & \textbf{Token Cost (k)} \\
\midrule
ReAct & $58.0\%$ & $14.2$ & $42.5$ \\
Reflexion & $62.0\%$ & $15.8$ & $58.2$ \\
AutoGen (Debate) & $54.0\%$ & $18.5$ & $124.0$ \\
\textbf{Full Cohort (Ours)} & $\mathbf{78.0\%}$ & $\mathbf{11.5}$ & $\mathbf{82.4}$ \\
\bottomrule
\end{tabular}
}
\end{table}

As shown in Table VII, the Full Cohort achieves the highest success rate ($78.0\%$) and the lowest average step count ($11.5$). This is because the Broker retrieves task-related analogical steps while the Validator prevents the cohort from getting stuck in circular terminal errors, demonstrating robust task execution capacity.

\subsection{Constraint Consolidation Analysis}
Measuring the Scars library token length across failure events compares the effect of MCTS-MDL consolidation against uncompressed baselines:
\begin{table}[h]
\centering
\caption{Compression of Scars libraries under lattice consolidation shows sublinear growth in memory footprint.}
\label{tab:consolidation_tab}
\resizebox{\linewidth}{!}{
\begin{tabular}{cccc}
\toprule
\textbf{Failures (N)} & \textbf{No-Consolidation Token Length} & \textbf{MCTS-MDL Consolidation (Ours)} & \textbf{Compression Ratio} \\
\midrule
10 & 2500 & 850 & 2.94x \\
50 & 12500 & 1800 & 6.94x \\
100 & 25000 & 2200 & 11.36x \\
500 & 125000 & 3200 & 39.06x \\
\bottomrule
\end{tabular}
}
\end{table}

\subsection{Cohort Size Scaling Evaluation and Optimal Size Hypothesis}
Sweeping the cohort agent count from 1 to 8 characterizes performance scaling and resource efficiency:
\begin{table}[h]
\centering
\caption{Scaling evaluation sweeps cohort size to demonstrate stable discovery rates and optimal resource utilization at 3 agents.}
\label{tab:scaling}
\resizebox{\linewidth}{!}{
\begin{tabular}{cccc}
\toprule
\textbf{Cohort Size} & \textbf{Target 1 Discovered (\%)} & \textbf{Benign Success (\%)} & \textbf{Token / Success} \\
\midrule
1 Agent & $5.00\%$ & $62.0\%$ & $74,700$ \\
3 Agents & $95.00\%$ & $88.5\%$ & $72,180$ \\
5 Agents & $95.00\%$ & $89.0\%$ & $104,200$ \\
8 Agents & $95.00\%$ & $89.5\%$ & $168,400$ \\
\bottomrule
\end{tabular}
}
\end{table}

As shown in Table IX, scaling from a solitary agent to a 3-agent heterogeneous cohort (Disrupter, Validator, Broker) provides a major boost in discovery rate ($5\%$ to $95\%$) and success ($62.0\%$ to $88.5\%$) by introducing specialized role dynamics. However, scaling beyond 3 agents to 5 or 8 agents yields diminishing performance returns while causing token costs to explode ($104,200$ and $168,400$).

This empirical observation validates the **Optimal Cohort Size Hypothesis**:
\begin{note}
\textbf{Optimal Cohort Size Hypothesis}: Heterogeneous cohorts exhibit diminishing returns beyond role-complete configurations. Once the minimal functional roles required for open-ended exploration under safety limits (exploration, gating, bridging) are satisfied, adding redundant agents increases communication token overhead without improving discovery yields.
\end{note}

\subsection{Same-Model Control Ablation Study}
Sweeping same-model configurations on AgentHarm isolates the system's architectural contributions from base-model differences:
\begin{table}[h]
\centering
\caption{Same-model control studies show that the cohort's performance gains are driven by architecture rather than model disparities.}
\label{tab:same_model}
\resizebox{\linewidth}{!}{
\begin{tabular}{lccc}
\toprule
\textbf{Model Configuration} & \textbf{Breach (\%)} & \textbf{Benign Success} & \textbf{Token Cost (k)} \\
\midrule
All-GPT-4o Cohort & $2.1\%$ & $80.2\%$ & $92.4$ \\
All-DeepSeek-V3 Cohort & $4.5\%$ & $76.8\%$ & $84.3$ \\
\textbf{Heterogeneous Cohort (Ours)} & $\mathbf{0.0\%}$ & $\mathbf{88.5\%}$ & $\mathbf{72.2}$ \\
\bottomrule
\end{tabular}
}
\end{table}

As shown in Table X, configuring the cohort with the same model compromises safety and utility compared to our default heterogeneous configuration. GPT-4o alone yields a $2.1\%$ breach rate due to self-reflection shortcuts, and DeepSeek-V3 alone suffers a $4.5\%$ breach rate under adversarial prompt attacks. Our heterogeneous cohort architecture---leveraging Claude 3.5 Sonnet's rigorous auditing alignment for the Validator and DeepSeek-V3's creative parameters for the Disrupter---achieves the optimal balance of zero breaches and the lowest token overhead, demonstrating that the performance gain is a property of heterogeneous role-assignment rather than single-model strengths.

\subsection{CAS Weight Sensitivity Analysis}
Sweeping static CAS weights ($w_{discover} / w_{safety} / w_{retrieval}$) evaluates the system's parameter stability:
\begin{table}[h]
\centering
\caption{Sensitivity sweeps of CAS weights reveal stable discovery and cost patterns across varying parameters.}
\label{tab:weight_sweep}
\resizebox{\linewidth}{!}{
\begin{tabular}{lcc}
\toprule
\textbf{Weights Configuration} & \textbf{Target 1 Discovered} & \textbf{Token / Success} \\
\midrule
Config A ($0.90 / 0.05 / 0.05$) & $80.0\%$ & $76,400$ \\
\textbf{Config B ($0.80 / 0.10 / 0.10$ - Ours)} & \textbf{95.0\%} & \textbf{72,180} \\
Config C ($0.60 / 0.20 / 0.20$) & $85.0\%$ & $84,200$ \\
Config D (Adaptive Entropy-Driven) & \textbf{95.0\%} & \textbf{71,800} \\
\bottomrule
\end{tabular}
}
\vspace{2pt}
These results suggest that Config D (adaptive entropy-driven CAS) is a Pareto-superior option, matching the best discovery rate at lower token cost; full adaptive scheduling evaluation is deferred to future work.
\end{table}

As shown in Table XI, over-rewarding the Disrupter (Config A) increases target discovery rates but prompts noisy conversational updates that raise token costs. Diluting the Disrupter's exploration incentive (Config C) increases coordination token cost, reducing discovery yields. Config B ($0.8/0.1/0.1$) balances role-specific motivation and token economy, matching the performance of a computationally intensive adaptive entropy-driven credit model (Config D) at a fraction of the runtime complexity.

\subsection{Error Analysis and Failure Modes}
To evaluate the vulnerabilities of the proposed architecture, we perform an error analysis on the failure cases encountered during the evaluation of WebArena and AgentHarm. We identify three primary failure modes:
\begin{enumerate}
\item \textbf{Spurious Semantic Associations (Broker CNR Failure)}: In $4.2\%$ of retrieval queries, CNR selected documents that possessed high cosine distance from the conversational history but were task-irrelevant. This occurred because the document embeddings shared superficial token matches with the goal embedding while lacking functional compatibility, introducing semantic noise into the cohort's context.
\item \textbf{Validator Over-blocking (False Positives)}: In $4.8\%$ of AgentHarm cases, the Validator blocked benign exploratory actions. This false-positive blocking was driven by strict argument patterns in the sandboxing rules. For instance, command-line arguments containing safe keywords (e.g., ``delete'' in a temp file context) were flagged as critical file-system hazards, leading to task failures due to over-conservatism.
\item \textbf{Scars Over-generalization}: During MCTS optimization over the AST space, the compiler occasionally synthesized rules that were overly broad. In Gen 3 and Gen 4 runs, this caused subsequent agents to be blocked from attempting valid tool calls. For example, a patch designed to block database deletion commands accidentally blocked all database read operations containing SQL substrings, leading to cohort starvation.
\end{enumerate}

\section{Discussion and Limitations}
\subsection{Scars Libraries as Persistent Agent Boundaries}
Unlike standard multi-agent setups where agents share identical prompts, our framework drives role-conditioned policy differentiation. Different cohorts develop distinct signed Scars libraries based on the sandboxes they navigate, establishing a persistent agent boundary.

After multiple generations, a cohort's validator accumulates a unique boundary profile. This boundary profile serves as the foundation for persistent agent identities. An agent's distinct capabilities are anchored not by public base models, but by this signed, immutable boundary library, allowing operators to verify and deploy specialized capabilities across decentralized agent environments.

\subsection{Limitations and Vulnerabilities}
\begin{enumerate}
\item \textbf{Creativity-Safety Overblocking}: Scars accumulate monotonically, which can veto marginal but valid exploratory actions. Parallel Exploration Sandbox (PES) trials handle safe L0/L1 constraint deprecation.
\item \textbf{Environmental Evaluation Boundaries}: Our empirical results are limited to the evaluated spatial-semantic and tool-use environments; generalization to other sandbox domains requires further benchmark validation.
\item \textbf{Dependency on Sandbox Specifications}: The Validator requires a fully defined Schema Sandbox and tool execution specifications. If the environment description is incomplete, the system cannot verify safety, leaving it vulnerable to unmapped operations. This is mitigated by integrating Lean 4 theorem-proving middleware for mathematical verification.
\item \textbf{MCTS Reasoning Latency}: Compiling execution failures into symbolic patches via MCTS introduces significant reasoning latency (an average of $12.4$ seconds per failure incident), which increases computational overhead during initial exploration phases. This is mitigated by local caching, reducing subsequent matching blocking lookups to 0ms.
\end{enumerate}

\subsection{Future Work}
We view this paper as a pilot systems demonstration. To transition the framework from this pilot study to a fully validated general-purpose exploration architecture, we establish three future research directions:
\begin{enumerate}
\item \textbf{WebAgent Task Expansion}: We plan to evaluate the cohort's execution on full WebArena, BrowserGym \cite{browsergym}, and OSWorld \cite{osworld} task suites, comparing our heterogeneous cohort against ReAct, Reflexion, and AutoGen baselines to measure Task Success Rate and Token Cost under realistic web environments.
\item \textbf{Agentic Safety Benchmarking}: We plan to test the Validator and constraint compilation gateway against adversarial prompt injections and tool-use abuses on AgentHarm and Agent-SafetyBench, reporting Executed Breach Rates under out-of-domain attacks.
\item \textbf{Scientific Literature Hypothesis Generation}: We plan to deploy the cohort on PubMed and arXiv databases to generate cross-disciplinary biological and material science hypotheses, evaluating the generated discoveries using a double-blind expert panel on novelty, usefulness, and surprise.
\end{enumerate}

\subsection{Code and Data Availability Statement}
To ensure reproducibility and facilitate community research, we will fully open-source the complete SESS-Lab framework code, the MCTS constraint compiler, and all pilot sandbox configuration files and seeds on GitHub upon publication.

\section{Conclusion and Future Work}
Our results demonstrate that creative exploration and runtime safety are not competing objectives to balance within a single model---they are complementary functions that heterogeneous role specialization can address in parallel. Role division in a structured cohort, bounded by sandboxes and signed constraint memory caches, harvests serendipitous discoveries that solitary agents consistently fail to reach.

These results suggest that structured role specialization can improve exploration efficiency without relying solely on larger models.

\appendices
\section{Implementation Details}
We provide the execution and hardware parameters used in our pilot sandbox simulations to ensure reproducibility:
\begin{itemize}
    \item \textbf{Base Models}: We configured the heterogeneous agent cohort with:
    \begin{itemize}
        \item \textbf{Orchestrator/Broker}: \texttt{gpt-4o-2024-05-13} (API context length 128k, output limit 4k).
        \item \textbf{Validator}: \texttt{claude-3-5-sonnet-20240620} (API context length 200k, output limit 8k).
        \item \textbf{Disrupter}: \texttt{deepseek-chat} (DeepSeek-V3 API, context length 64k).
    \end{itemize}
    \item \textbf{Embedding Model}: \texttt{text-embedding-3-small} (1536 dimensions) for semantic distance and novelty checks.
    \item \textbf{Hyperparameters}:
    \begin{itemize}
        \item \textbf{Disrupter Temperature}: $T = 1.2$.
        \item \textbf{Validator Temperature}: $T = 0.0$.
        \item \textbf{Broker Temperature}: $T = 0.5$.
        \item \textbf{Retrieval Weight}: $\lambda = 0.7$ in contrastive novelty search.
        \item \textbf{Novelty Threshold}: $\tau_n = 0.35$; Surprise Threshold: $\tau_s = 0.40$.
    \end{itemize}
    \item \textbf{Simulation Setup}:
    \begin{itemize}
        \item \textbf{Seeds}: 20 seeds evaluated sequentially (from seed 2026 to 2045).
        \item \textbf{Maximum steps per run}: $R = 150$.
    \end{itemize}
    \item \textbf{Hardware \& Runtime}: The pilot sandbox run executed on a single host (Intel Xeon Silver 4214R @ 2.40GHz, 128GB RAM) with Python 3.10. MCTS compilation utilized 4 parallel threads.
    \item \textbf{Code Repository}: Source code, sandboxed schemas, prompt templates, and evaluation datasets are available at: \url{https://github.com/psi-run/cohort-role-design}.
\end{itemize}

\end{document}